\documentclass[runningheads]{llncs}
\usepackage{graphicx}
\usepackage{amsfonts,amsmath,amssymb}
\usepackage{mathtext}
\usepackage[ruled,vlined]{algorithm2e}
\begin{document}
\title{Braid-based architecture search\thanks{The computing resources of the Shared Facility Center "Data Center of FEB RAS" (Khabarovsk) were used to carry out calculations.}}
\titlerunning{Braid-based architecture search}
%
\author{Olga Lukyanova\orcidID{0000-0001-6864-9875} \and
Oleg Nikitin\orcidID{0000-0001-9139-5180} \and
Alex Kunin\orcidID{0000-0002-5221-0590}}
\authorrunning{O. Lukyanova et al.}
%
\institute{Computing Center, Far Eastern Branch of the Russian Academy of Sciences, Khabarovsk, 680000, Russia}
\maketitle              
\begin{abstract}
In this article, we propose the approach to structure optimization of neural networks, based on the braid theory. The paper describes the basics of braid theory as applied to the description of graph structures of neural networks. It is shown how networks of various topologies can be built using braid structures between layers of neural networks. The operation of a neural network based on the braid theory is compared with a homogeneous deep neural network and a network with random intersections between layers that do not correspond to the ordering of the braids. Results are obtained showing the advantage of braid-based networks over comparable architectures in classification problems.

\end{abstract}
\section{Procedural generation of neural networks} The architecture of neural networks is selected by studying their accuracy and ability of generalization. This approach is not optimal and requires a lot of time and computational resources. So, it can be useful to replace it by the automatic optimization of neural network architectures. Automatic approaches imply the use of algorithmic (procedural) methods for the generation of neural networks, that is, the application of rules and procedures that create certain \cite{1} sequences. This approach will be useful for generating deep neural network architectures, since the optimal setting of their structure is nontrivial and directly depends on the problem being solved. Algorithmic search of neural network architectures is an actively developing research topic of current interest \cite{2}.

Usually, experts select the structures and hyperparameters of deep neural networks manually. This can lead to highly efficient specialized architectures. The disadvantage of this method is the need for long-lasting experiments to solve each specific problem. To overcome this, the Neural Architecture Search (NAS) methods have been proposed, such as NAS-RL \cite{7} and MetaQNN \cite{8}, which use reinforcement learning, a method of machine learning, in which a model that does not have information about the system but has the ability to perform any actions in it to optimize a parameter is trained. NAS-RL and MetaQNN require a lot of computation in the selection process to assess the effectiveness of the resulting architecture. These algorithms imply the direct setting of the network modules' structure in the optimization process.

NAS approaches usually involve direct specification of optimized architectures, which often expands the search space and complicates the optimization. Procedural generation of architectures complements these approaches and involves the creation of architectures based on algorithmic rules. Such approaches in evolutionary optimization are called indirect coding -- the generation rule is encoded, not the direct result. Examples of such method are: Convolutional Neural fabrics (CNF) \cite{10}, PathNet \cite{11}, and Budgeted Super Networks (BSN) \cite{12}. All these approaches are united by the method of forming architectures, which implies the generation of types of modules and connections between them based on algorithmic approaches.

In general, all of the compared approaches require complex computations to optimize the network topology, and they do not support multi-column and parallel architectures. Many of them involve one or the other of the brute force methods. To build optimal architectures, there are not enough analytical estimates of information indicators of network performance and, nowadays, the selection is made empirically. This makes the problems of procedural generation of neural architectures computationally complex or suboptimal, therefore, research in the field of neural network topologies and the search for theoretical estimates of information dynamics in neural networks is an important task.

To investigate such a procedure, it is proposed to use a low-dimensional topology. Braid theory as part of low-dimensional topology allows to systemize and order the interconnections of computational graphs to overcome the the disadvantages of previous approaches since it eliminates the need for brute force. 

In this article, the authors propose the use of modern mathematical methods to optimize the structure of neural networks. It also provides an overview of the application of braid theory to improve the performance of neural networks. In previous research \cite{5} procedural generation techniques based on matrix filters and braid theory were studied by authors. In the present article, we modify such an approach to generate and optimize BraidNets.

\section{BraidNet deep neural network architecture} 
\begin{figure}[!htb]
\centering 
\includegraphics[width=50mm]{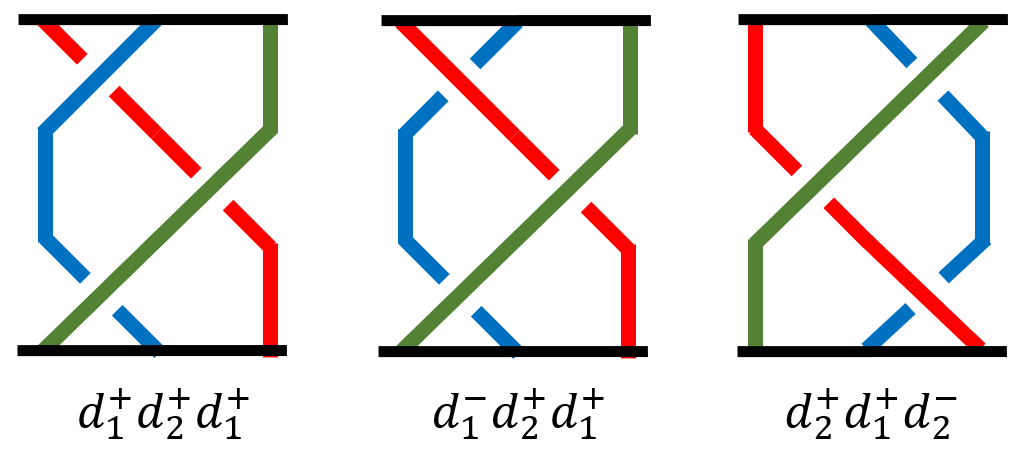}
\caption{The example of three braids and their letter descriptions.} 
\label{img:2}  
\end{figure}

Braid theory is a section of topology and algebra that studies braids (pairwise intersections of strands) and braid groups composed of their equivalence classes. A mathematical braid consists of $m$ strands (that is, curves in space) that start at $m$ points on a horizontal line and end at $m$ points on another horizontal line below \cite{25}. Examples of three braids are shown in Figure~\ref{img:2}.

The braid can be represented as a "braid word", i.e. the product of elements  $d_i^{\pm}$, where $d_i$ are the braid diagrams on three strands. Thus, the elements $d_i,...,d_{n-1}$ generate the group $D_n$. Let us denote by $d_i$ a braid of $m$ strands, the $i$-th strand of which passes "under" $(i+1)$-th, and the other strands have no crossings. This is a sequence of notation $d_1^+$, $d_1^-$, $d_2^+$, $d_2^-$ and so on. It can be written which strand of the braid goes under or above the other strand. A braid diagram with $m$ strands has positions from 1 to $m$. If the strand $m$ is crossed from above by the strand $m+1$, it is written as $d_m^-$. And vice versa: $d_m^+$ for the case when the strand $m+1$ passes under the strand $m$. Thus, you can "move" the strands of the braid, but you cannot disconnect them from the points at which they begin and end, cut and glue them. Various versions of braids that arise after these movements are equivalent (isotopic) to the original braid \cite{27}.

The structures of neural networks set by the braids allow the calculation and comparison of weights in network modules. Thus, braid theory can be applied to automatically form and dynamically change the structure of neural network. This allows avoiding manual iteration over the sequences of comparisons between modules. The network parameters are set by the number of layers and the number of strands of the braid, where each of the strands solves the problem of finding its own data class among all input examples ("one against all" problem). The number of strands corresponds to the number of classes. Further, all possible existing options for braids are determined for the given parameters, that is, the number of strands and their length. The structure of the neural network (braid words) is randomly selected.

The input data is fed to each of the braid strands at the same time. After that, we check if there any intersection: if it exists, weights of the modules whose strands participate in the intersection are recalculated (see below for more details). The data of these modules is transmitted further over the network, and then, the residual of the prediction from the fact is calculated separately for each of the strands and the training is performed using the backpropagation method. The procedure is repeated for all training data.

\begin{figure}[!htb]
\centering
\includegraphics[width=50mm]{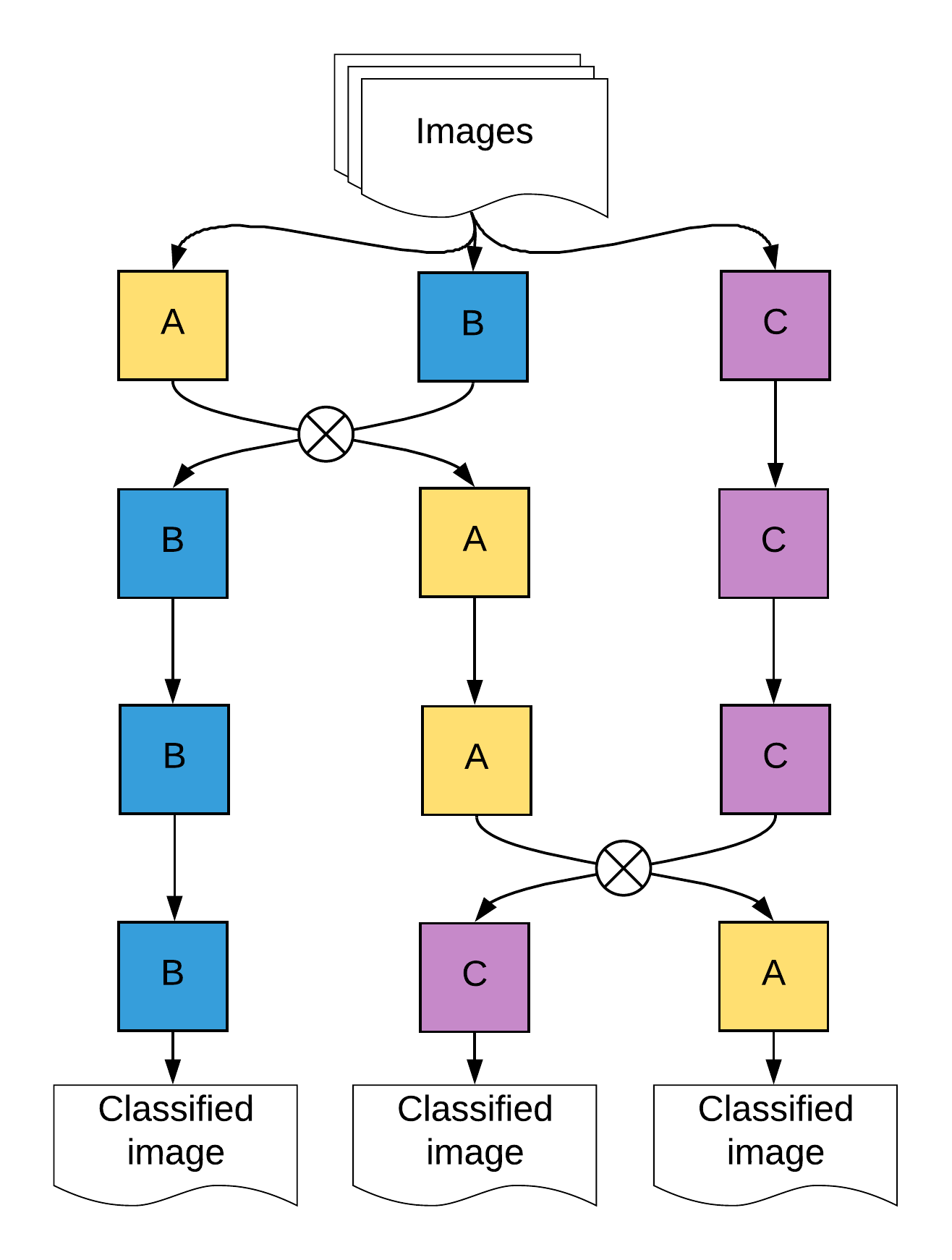}
\caption{An example of a neural network architecture BraidNet, based on the braid theory, consisting of three braid strands.}
\label{img:11}
\end{figure}

An example of the resulting architecture of the deep neural network is shown in Figure~\ref{img:11}. This network consists of braid strands, the number of which is determined by the number of classes recognized by the network. Each strand learns to recognize the class assigned to it.

The combination of structures defined by braids and dynamic modification of the computation graph makes it possible to obtain structures that naturally allow solving problems of multiclass classification, since the presence of many strands in the network helps to learn individual strands for each class, and mix and evaluate weight information -- to combine some common features for different classes.

Interconnection structures between neural network modules, defined using braid theory, can be used to build network topologies with multiple strands. The architecture of this neural networks allows to transmit information between different classes, where one class belong to the one strand of the braid. Then, strands with the network modules, for example, $A$ and $B$, solve the problem of finding their classes among all classes $N$. Below, we mix weights between the layers to share information between the classes. Intersections between modules are specified using braid theory. The transmission structure between the layers at the intersection points is selected based on the location of the strand of specific class. It depends on if strand of one class goes above or under the other class strand. Then the weights of both classes are recalculated as follows:

\begin{equation}
\label{eq:eq1}
\begin{cases}
W^A_{L+1} = W^A_{L} \\
W^B_{L+1} = W^B_{L} + \alpha W^A_{L}
\end{cases}
\end{equation}
where  \\
$A$ -- the module and the strand that goes under the other strand $B$, \\
$B$ -- the module and the strand that goes above the other strand $A$, \\
$A,B\in N$, \\
$N$ -- the number of strands (number of classes), \\
$W^A_{L}$, $W^B_{L}$ -- the weights of the modules $A$ and $B$ on the layer $L$, \\
$W^A_{L+1}$, $W^B_{L+1}$ -- the weights of the modules $A$ and $B$ on the layer $L+1$, \\
$\alpha$ -- the predetermined influence coefficient. \\

\begin{figure}[!htb]
\centering
\includegraphics[width=70mm]{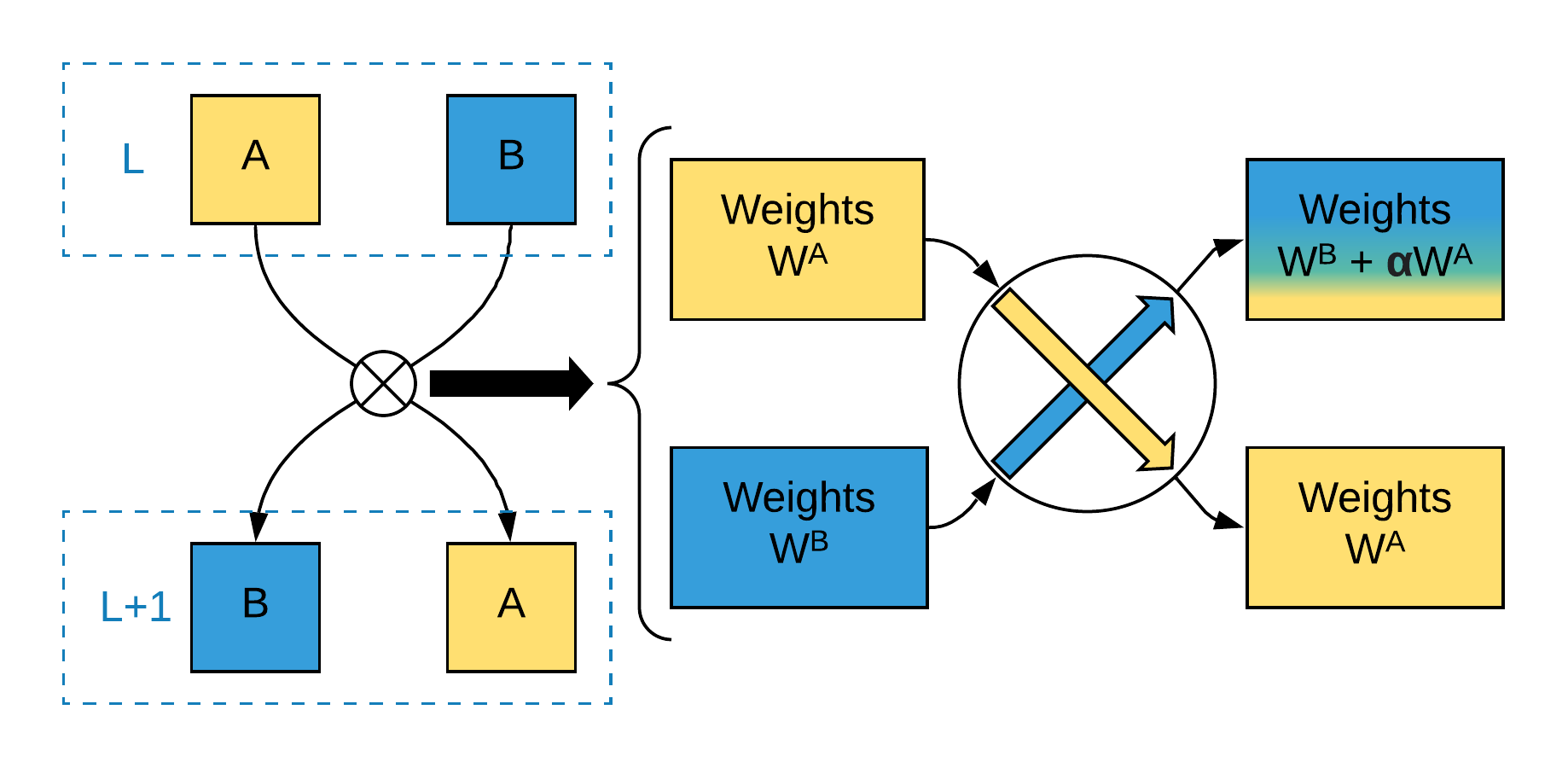}
\caption{An example of weights' mixing between network layers} 
\label{img:5}  
\end{figure}

If strands will be reversed crossed then the weights will be calculated likewise in reverse. The scheme of information transfer between the network layers at the strands' crossings for such a case will correspond to the figure~\ref{img:5}.

Thus, the calculation of the weights of modules with crossed strands in the process of training neural network allows to dynamically change the paths of information transfer. The modules collect additional useful information about the classes of nearby modules that increase the speed and accuracy of classification. This kind of comparison is possible between modules only in pairs, which well complements the pairwise character of intersections in braid structures described above.

\section{Simulation results} 
\begin{figure}[!htb]
\centering
\includegraphics[width=120mm]{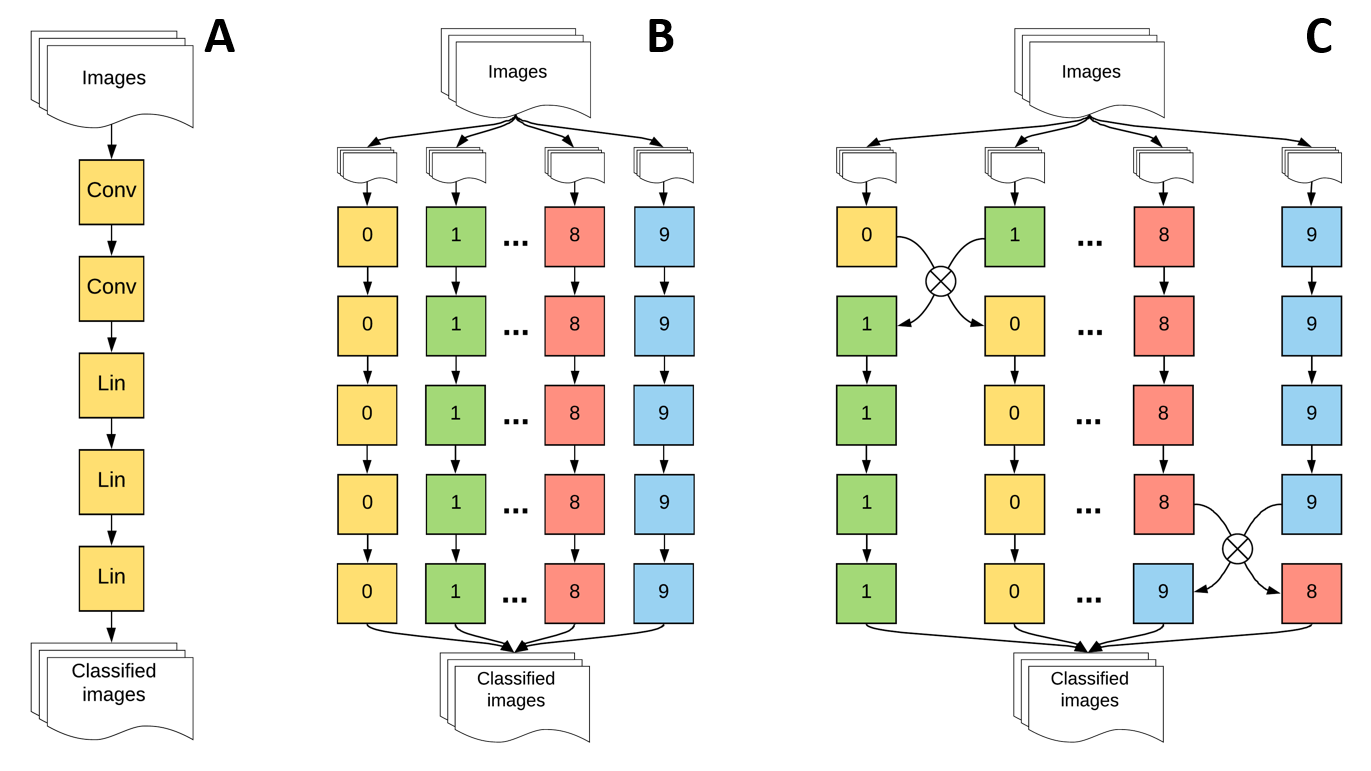}
\caption{The DNN (A), 10DNN (B) and BraidNet (C) architectures}
\label{img:6}
\end{figure}

To assess the performance of the BraidNet and compare it with other architectures in the context of multiclass classification problems, the neural networks were applied to classical image recognition problem -- SVHN. The SVHN recognizes the numbers on the house number indicators. The BraidNet was implemented with crossings between strands specified by the braid. For comparison, we chose a deep neural network (DNN) and 10DNN -- a deep neural network with 10 parallel strands (see Figure~\ref{img:6}). 

All parameters on the layers of the compared networks were the same. Thus, to solve SVHN classification problem, the BraidNet architecture was used, each of ten strands consisted of five modules. The first two modules contained convolutional layers with a kernel equal to 5, and activation layers and pooling with a step equal to 2. The next two modules consisted of fully connected layers. At the end of each strand, a ReLU (Rectified Linear Unit) activation function was applied.

\begin{figure}[!htb]
\centering
\includegraphics[width=90mm]{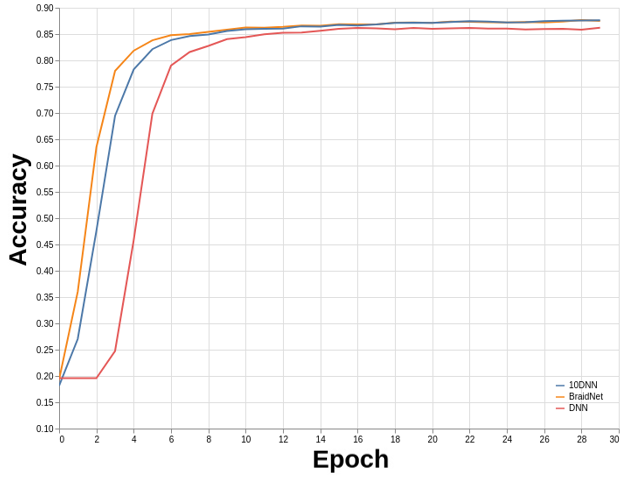}
\caption{The results of modeling the SVHN problem for DNN, 10DNN and BraidNet architectures}
\label{img:7}
\end{figure}

The learning results for SVHN numbers recognition problem are shown in the figure~\ref{img:7}. The BraidNet architecture allowed to quickly achieve an increase in accuracy than 10DNN, and after 6th epoch it stayed close to 10DNN, while DNN had much worse results. The figure shows that networks based on the BraidNet architecture are able to learn faster than deep neural networks, even the DNN with 10 parallel networks (strands). Thus, architecture with crossings between the strands outperforms architectures without crossings, like DNN or 10DNN, because of weight mixing between the classes.

\begin{figure}[!htb]
\centering
\includegraphics[width=80mm]{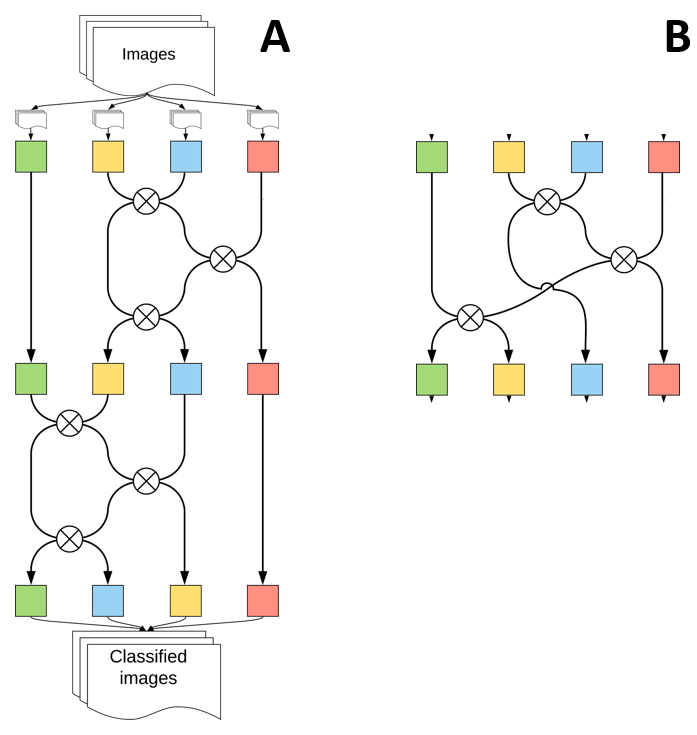}
\caption{(A) BraidNet with braids between the layers, (B) neural network with random crossings between layers}
\label{img:8}
\end{figure}

However, to understand how BraidNet works we build even more complex BraidNet architecture, so it consists not only crossings between modules, but whole braids between the layers (see Figure~\ref{img:8}A). 

\begin{figure}[!htb]
\centering
\includegraphics[width=90mm]{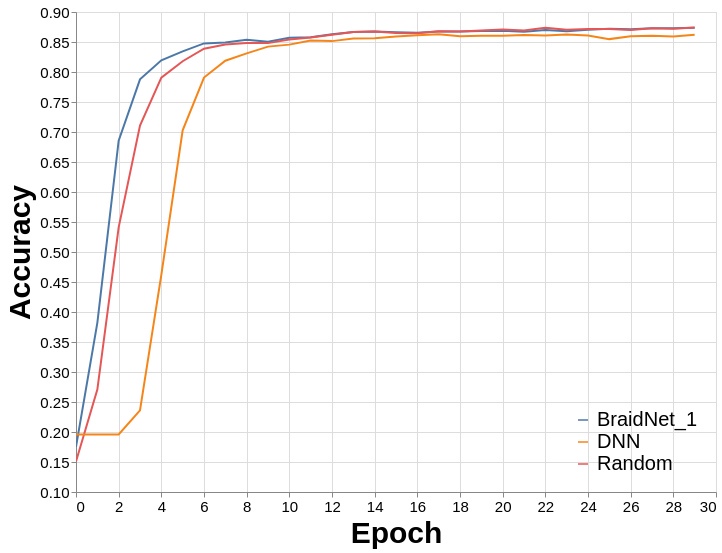}
\caption{The results of modeling the SVHN problem for DNN, Random and BraidNet architectures}
\label{img:9}
\end{figure}

And, to compare new BraidNet architecture and test wether it's advantage is braid crossings or it could be any random crossings we build another network architecture -- Random, similar to BraidNet, but with random strands' crossings between layers (see Figure~\ref{img:8}B). So, it is not keep the braid theory rules and could have direct connections with distant strand. The results of modeling the SVHN problem for DNN, Random and BraidNet architectures are shown in the figure~\ref{img:9}.

It can be seen, that BraidNet architecture again shows the best results at the first six epochs, and after that it stays as good as Random architecture. Thus, nevertheless, the architectures with braid rules and with random crossings are remain the same at the last epochs, the BraidNet still learns faster.

\begin{figure}[!htb]
\centering
\includegraphics[width=120mm]{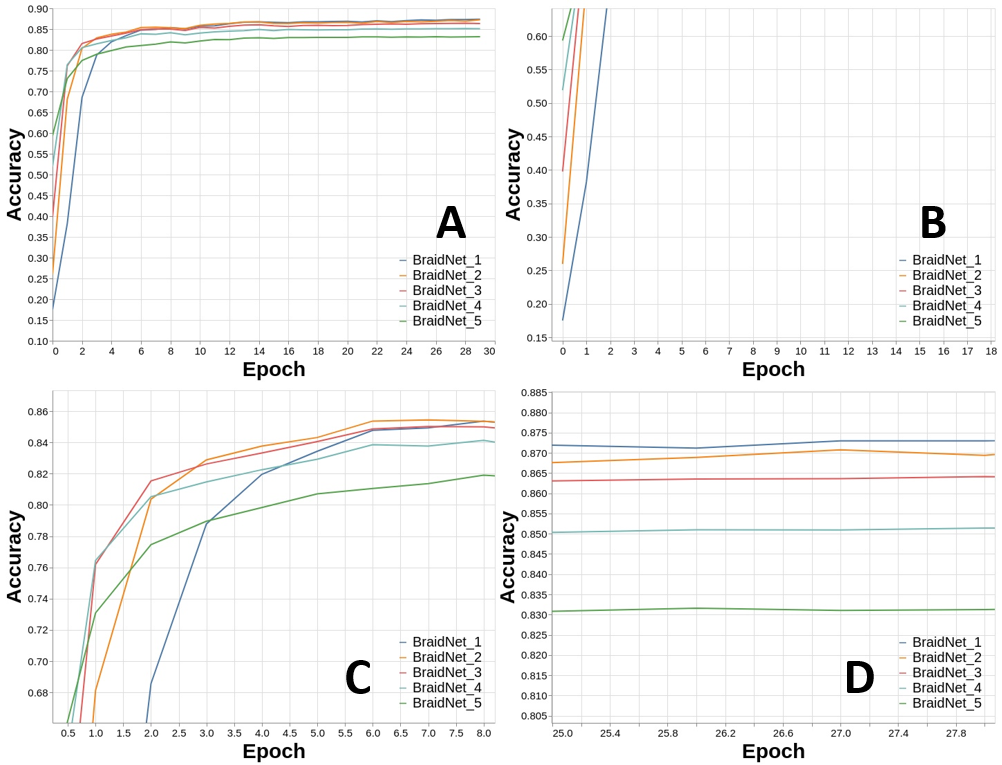}
\caption{The accuracy results of modeling the SVHN problem for BraidNet architectures with 1, 2, 3, 4, 5 braid strands' crossings between layers}
\label{img:10}
\end{figure}

To verify how the number of braid strands' crossings affects on the classification results, we compared BraidNet architectures (Figure~\ref{img:8}A) with one, two, three, four and five braid strands' crossings between the layers respectively. The results of this comparison are shown in figure~\ref{img:10}.

The figure~\ref{img:10}A shows an accuracy for all 30 epochs of modeling, and figures~\ref{img:10}B--\ref{img:10}D show beginning, middle and end close-up images of figure~\ref{img:10}A. It can be seen, that at the beginning BraidNet with 5 strands' crossings between the layers is showing much better results than other BraidNets. But, in the middle of modeling, it is becoming the worst by the accuracy results. At the end, around 25-30 epochs BraidNet with 1 crossing between the layers is showing the best accuracy despite previous the worst result from all 5 architectures. Thus, indeed, the number of braid strands' crossings between the layers of BraidNet architecture correlates with the accuracy of training. But, at the first epochs it is better to use large number of crossings, and from the middle of training, when the training stars to slow, it is preferable to reduce numer of crossings. 

Thus, the simulation results showed the comparative advantage of BraidNet in learning speed and classification accuracy. The BraidNet architecture is a promising algorithm for further application and research in complex problems of pattern recognition.

\section{Conclusion} As part of research on the automatic generation of neural network structures, the authors proposed and studied the BraidNet approach, which includes the generation of network topology using braid theory. The work also implemented a number of simplified architectures to analyze the performance of the BraidNet architecture. The neural networks, such as BraidNet, DNN, 10DNN were applied to solve the SVHN multiclass image classification problem. The simulation results showed the comparative advantage of BraidNet in learning speed and classification accuracy. In general, BraidNet training is less computationally intensive than similar approaches.

It should be noted that the study did not carry out additional optimization of the BraidNet network hyperparameters, such as the learning rate and dropout rate, and did not conduct experiments to introduce noise into the training set. These actions, as well as the inclusion of various layers in the network, except for convolutional ones (pooling, dropout, ReLu, etc.), and optimization number of strands' crossings between layers can increase the final accuracy of the network. Despite the existing opportunities for improving the BraidNet network, it can be applied in practice in the form obtained during the implementation. Thus, the BraidNet approach can be used to optimize neural networks when solving classification problems.

\bibliographystyle{splncs04}
\bibliography{mybibliography}

\end{document}